\documentclass[11pt]{article}

\usepackage[final]{coling}

\usepackage{times}
\usepackage{latexsym}

\usepackage[T1]{fontenc}

\usepackage[utf8]{inputenc}

\usepackage{microtype}

\usepackage{inconsolata}

\newcommand*\samethanks[1][\value{footnote}]{\footnotemark[#1]}
\usepackage[ruled,linesnumbered]{algorithm2e}
\usepackage{amsmath}
\usepackage{amsthm}
\usepackage{enumitem}
\setlist{left=\parindent} 
\usepackage{xcolor}
\usepackage{graphicx}
\usepackage{booktabs}
\usepackage{multirow}
\usepackage{xspace}
\usepackage{balance}
\usepackage{bbold}
\usepackage{pifont}
\usepackage{colortbl}
\usepackage[normalem]{ulem}
\useunder{\uline}{\ul}{}
\usepackage{tikz}

\theoremstyle{definition}

\newcommand{\ie}{\emph{i.e.,}\xspace}

\newcommand{\wrt}{\emph{w.r.t.}\xspace}

\newcommand{\stitle}[1]{\noindent{\bf #1}\hspace{1.5pt}}

\newcommand{\sys}{\textsc{ComEM}\xspace}

\everydisplay{\textstyle}

\newcommand{\bminus}{
  \begin{tikzpicture}[baseline=0.0ex,line width=1.5,scale=0.12]
    \draw (0,1) -- (2,1);
  \end{tikzpicture}
}
\newcommand{\bplus}{
  \makebox[2.5pt][c]{
    \begin{tikzpicture}[baseline=0.0ex,line width=1.3,scale=0.12]
      \draw (0,1) -- (2,1);
      \draw (1,0) -- (1,2);
    \end{tikzpicture}
  }
}

\usepackage{etoolbox}
\makeatletter
\patchcmd{\@setref}{\bfseries ??}{\bfseries \color{red} ??}{}{}
\patchcmd{\NAT@citex}{\bfseries ?}{\bfseries \color{red} ?}{}{}
\patchcmd{\NAT@citexnum}{\bfseries ?}{\bfseries \color{red} ?}{}{}
\ifdefined\HyRef@autosetref
  \patchcmd{\HyRef@autosetref}{\bfseries ??}{{\bfseries \color{red} ??}}{}{}
\fi
\makeatother

\title{Match, Compare, or Select? An Investigation of Large Language Models for Entity Matching}

\author{
  Tianshu Wang$^{1,2}$,
  Xiaoyang Chen$^3$,
  Hongyu Lin$^1$,
  Xuanang Chen$^1$\thanks{Corresponding author.}, \\
  {\bf
  Xianpei Han$^{1,4}$\samethanks,
  Hao Wang$^5$,
  Zhenyu Zeng$^5$,
  Le Sun$^{1,4}$} \\
  $^1$Chinese Information Processing Laboratory, Institute of Software, Chinese Academy of Sciences \\
  $^2$Hangzhou Institute for Advanced Study, University of Chinese Academy of Sciences \\
  $^3$University of Chinese Academy of Sciences \\
  $^4$State Key Laboratory of Computer Science, Institute of Software, Chinese Academy of Sciences \\
  $^5$Alibaba Cloud Intelligence Group \\
  \{tianshu2020, hongyu, chenxuanang, xianpei, sunle\}@iscas.ac.cn, \\
  chenxiaoyang19@mails.ucas.ac.cn,
  cashenry@126.com, zhenyu.zzy@alibaba-inc.com \\
}
\begin{document}
\maketitle

\begin{abstract}
  Entity matching (EM) is a critical step in entity resolution (ER).
  Recently, entity matching based on large language models (LLMs) has shown great promise.
  However, current LLM-based entity matching approaches typically follow a binary matching paradigm that ignores the global consistency among record relationships.
  In this paper, we investigate various methodologies for LLM-based entity matching that incorporate record interactions from different perspectives.
  Specifically, we comprehensively compare three representative strategies: matching, comparing, and selecting, and analyze their respective advantages and challenges in diverse scenarios.
  Based on our findings, we further design a compound entity matching framework (\sys) that leverages the composition of multiple strategies and LLMs.
  \sys benefits from the advantages of different sides and achieves improvements in both effectiveness and efficiency.
  Experimental results on 8 ER datasets and 10 LLMs verify the superiority of incorporating record interactions through the selecting strategy, as well as the further cost-effectiveness brought by \sys.
\end{abstract}


\section{Introduction}

\begin{figure}
  \centering
  \includegraphics[width=\columnwidth]{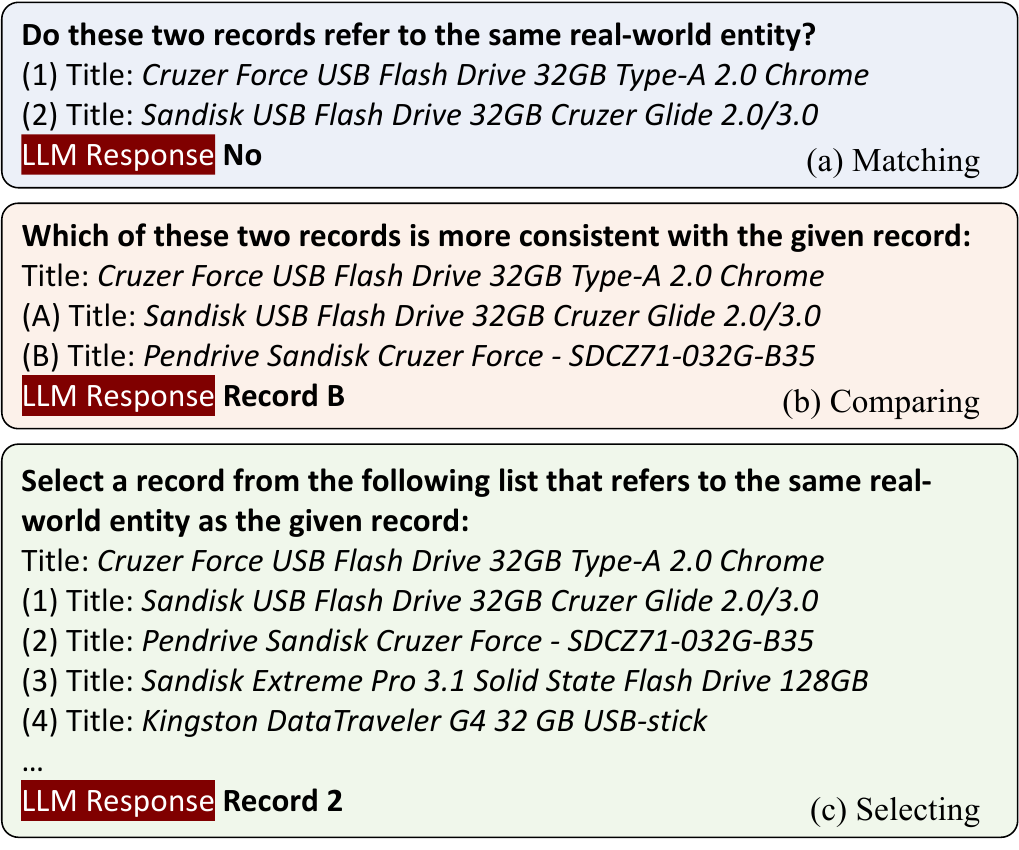}
  \caption{Three strategies for LLM-based entity matching.
    We omit other attributes of records for simplicity.}
  \label{fig:introduction}
\end{figure}

Entity resolution (ER), also known as record linkage~\citep{fellegi-69-theor-recor-linkag} or deduplication~\citep{DBLP:journals/tkde/ElmagarmidIV07}, aims to identify and canonicalize records that refer to the same real-world entity.
ER is a fundamental task of data integration and cleansing, with broad applications in maintaining data consistency, accurate data analysis, and informed decision making.
Entity matching (EM) serves as a critical step in entity resolution that uses complex techniques to identify matching records from potential matches filtered by the blocking step~\citep{DBLP:series/synthesis/2021Papadakis}.
The recent emergence of large language models (LLMs) has introduced a new zero- or few-shot paradigm to EM, showing great promise~\citep{DBLP:journals/pvldb/NarayanCOR22,DBLP:conf/adbis/PeetersB23,DBLP:conf/icde/FanHFC00024,DBLP:journals/corr/abs-2401-03426,DBLP:conf/edbt/PeetersSB25}.
LLM-based entity matching methods can achieve similar or even better performance than deep learning methods trained on large amounts of data, and are less susceptible to the unseen entity problem~\citep{DBLP:conf/ijcai/WangLFH0XCLZ22,DBLP:conf/edbt/PeetersDB24}.

However, current LLM-based entity matching methods identify matches by classifying each pair of records independently.
This \emph{matching} strategy ignores the global consistency\footnote{We refer to the interdependence of matching decisions in ER as global consistency. See \autoref{app:global_consistency} for more details.} among record relationships and thus leads to suboptimal results.
On the one hand, entity resolution requires more than independent classification due to the interconnected nature of record relationships~\citep{DBLP:journals/pvldb/GetoorM12}.
For example, in record linkage (\ie clean-clean ER), a single record from one data source typically matches at most one record from another data source, since there are usually no duplicates in a single database.
Unfortunately, matching-based approaches do not take advantage of this nature of record linkage.
On the other hand, this strategy ignores the capabilities of LLMs to handle multiple records simultaneously to distinguish similar records.
Using the records in \autoref{fig:introduction}(c) as an example, if ``Cruzer Glide'', ``Cruzer Force'', and ``Extreme Pro'' appear in different records of the same context, LLMs are more likely to recognize that they are different SanDisk flash drive models, which helps with accurate matching.
As a result, the \emph{matching} strategy cannot fully unleash the potential of LLMs in EM.

In this paper, we thoroughly investigate three strategies for LLM-based entity matching that incorporate record interactions from different perspectives, as shown in \autoref{fig:introduction}.
Specifically, apart from the conventional \emph{matching} strategy shown in \autoref{fig:introduction}(a), we investigate two additional strategies that leverage information from other records: 1) the \emph{comparing} strategy, which identifies the record out of two that is more likely to match the anchor record, as shown in \autoref{fig:introduction}(b); 2) the \emph{selecting} strategy, which directly chooses the record from a list that is most likely to match the anchor record, as shown in \autoref{fig:introduction}(c).
Our research suggests that for LLM-based entity matching, incorporating record interactions is critical and can significantly improve entity matching performance in various scenarios.
Among these strategies, the \emph{selecting} strategy is often the most cost-effective.
Nevertheless, we also observe that the selection accuracy varies significantly as the position of the matching record increases in the candidate list.
The position bias and limited long context understanding of current LLMs~\cite{DBLP:conf/acl/LevyJG24} hinder the generality of the \emph{selecting} strategy.

Based on our findings, we design a \emph{compound entity matching framework} (\sys) that leverages the composition of multiple strategies and LLMs.
Specifically, given an entity record and its $n$ potential matches obtained from the blocking step, we first preliminarily rank and filter these candidates using the local \emph{matching} or \emph{comparing} strategy, implemented with a medium-sized LLM.
We then perform fine-grained identification on only the top $k$ candidates using the global \emph{selecting} strategy, facilitated by a more powerful LLM.
This approach not only mitigates the challenges and biases faced by the \emph{selecting} strategy with too many options, but also reduces the cost of LLM invocations caused by composing multiple strategies.
Consequently, by integrating the advantages of different strategies and LLMs, \sys achieves a more effective and cost-efficient entity matching process.

To investigate different strategies and to evaluate \sys, we conducted in-depth experiments on eight ER datasets.
Experimental results verify the effectiveness of incorporating record interactions through the \emph{selecting} strategy, with an average 16.02\% improvement in F1 over the current \emph{matching} strategy.
In addition, we examined the effect of 10 different LLMs using these strategies on identification or ranking.
Ultimately, \sys is able to further improve average F1 of the single \emph{selecting} strategy by up to 4\% while reducing the cost.

\stitle{Contributions.} In general, our contributions can be summarized as follows\footnote{Our code is available at \href{https://github.com/tshu-w/ComEM}{github.com/tshu-w/ComEM} to facilitate reproduction of our results.}:
\begin{itemize}
  \item We investigate three strategies for LLM-based entity matching, and delve into their advantages and shortcomings in different scenarios.
  \item We design a \sys framework by integrating the advantages of different strategies and LLMs to address the challenges of EM.
  \item We conduct thorough experiments to investigate these strategies for EM and verify the effectiveness of our proposed framework.
\end{itemize}


\section{Related Work}

\begin{table*}
  \resizebox{\textwidth}{!}{%
    \begin{tabular}{@{}lllll@{}}
      \toprule
      \textbf{Records}                                                             & \textbf{Title}                                                & \textbf{Authors}                    & \textbf{Venue} & \textbf{Year} \\ \midrule
      Anchor                                                                       & Lineage Tracing for General Data Warehouse Transformations    & Yingwei Cui, Jennifer Widom         & VLDB           & 2001          \\ \midrule
      \multirow{4}{*}{\begin{tabular}[c]{@{}l@{}}Potential\\ Matches\end{tabular}} & Lineage Tracing for General Data Warehouse Transformations    & Yingwei Cui, Jennifer Widom         & VLDB Journal   & 2003          \\
                                                                                   & Tracing the lineage of view data in a warehousing environment & Yingwei Cui, Jennifer Widom, et al. & TODS           & 2000          \\
                                                                                   & Lineage tracing for general data warehouse transformations    & Y. Cui, J. Widom                    & VLDB           & 2001          \\
                                                                                   & … & … & … & … \\ \bottomrule
    \end{tabular}
  }
  \caption{Example of our formulation for entity matching: Given an anchor record, identify the matching record (if any) from its potential matches. This example is taken from the DBLP-ACM dataset.}
  \label{tab:example}
\end{table*}

\subsection{Entity Resolution}

Entity resolution has received extensive attention over the past decades~\citep{fellegi-69-theor-recor-linkag,DBLP:journals/pvldb/GetoorM12,DBLP:series/synthesis/2021Papadakis,Binette_2022}.
The blocking-and-matching pipeline has become the mainstream of entity resolution, where blocking filters out obviously dissimilar records and matching identifies duplicates through complex techniques.

\stitle{Blocking.} Traditional blocking approaches group records into blocks by shared signatures, followed by cleaning up unnecessary blocks and comparisons~\citep{DBLP:journals/corr/abs-2202-12521}.
Meta-blocking further reduces superfluous candidates by weighting potential record pairs and graph pruning~\citep{DBLP:journals/tkde/PapadakisKPN14}.
Recently, nearest-neighbor search techniques, especially cardinality-based ones, have gained more attention and achieved state-of-the-art (SOTA) results~\citep{DBLP:journals/pvldb/Thirumuruganathan21,DBLP:journals/pvldb/PaulsenGD23,DBLP:conf/naacl/WangZ24}.

\stitle{Entity Matching.} The open and complex nature of entity matching has spurred the development of various approaches to address this persistent challenge, including rule-based~\citep{DBLP:journals/vldb/BenjellounGMSWW09,DBLP:journals/tkde/LiLG15}, distance-based~\citep{DBLP:journals/expert/BilenkoMCRF03}, and probabilistic methods~\citep{fellegi-69-theor-recor-linkag,DBLP:conf/sigmod/WuCSCT20}, etc.
With the advent of deep learning methods~\citep{DBLP:conf/sigmod/MudgalLRDPKDAR18}, especially pre-trained language models (PLMs)~\citep{DBLP:journals/pvldb/0001LSDT20}, entity matching has made significant progress~\citep{DBLP:journals/tkdd/BarlaugG21,DBLP:journals/pacmmod/TuFTWL0JG23,DBLP:journals/pvldb/WuWDHZ23,DBLP:journals/corr/abs-2409-04073}.
The emergence of LLMs brings a new zero- or few-shot paradigm to entity matching~\citep{DBLP:journals/pvldb/NarayanCOR22,DBLP:journals/corr/abs-2410-12480,DBLP:conf/edbt/PeetersSB25}, alleviating training data requirements.
Most deep learning and LLM-based approaches treat entity matching as an independent binary classification problem, except for GNEM~\citep{DBLP:conf/www/ChenSZ21}, which models this task as a collective classification task on graphs.

\subsection{Large Language Model}

The advent of LLMs such as ChatGPT marks a significant advance in artificial intelligence, offering unprecedented natural language understanding and generation capabilities.
By scaling up the model and data size of PLMs, LLMs exhibit emergent abilities~\citep{DBLP:journals/tmlr/WeiTBRZBYBZMCHVLDF22} and can thus solve a variety of complex tasks by prompt engineering.
For more technical details on LLMs, we refer the reader to the related survey~\citep{DBLP:journals/corr/abs-2303-18223}.

While LLMs have shown promising results in classification and ranking tasks~\citep{DBLP:conf/emnlp/0001YMWRCYR23,DBLP:conf/naacl/QinJHZWYSLLMWB24}, applying LLMs to entity matching presents unique challenges and opportunities. Our work differs from previous research in three aspects: First, we propose a novel paradigm that formulates entity matching as a comparison or selection task. Second, we demonstrate that the effectiveness of pairwise and listwise strategies in entity matching exhibits different patterns compared to ranking. Finally, through a comprehensive cross-model and cross-strategy evaluation, we reveal several key insights about LLM-based EM, which motivate the design of our \sys framework.


\section{Entity Matching with LLMs}

In this section, we first present the problem formulation.
Then, we introduce three strategies for LLM-based entity matching.
Finally, we propose our \sys framework, which leverages the composition of multiple strategies and LLMs.

\subsection{Problem Formulation}

We formulate the task of entity matching as the process of identifying matching records from a given anchor record $r$ and its $n$ potential matches $R = \{r_1, r_2, \ldots, r_n \}$ obtained from blocking, as illustrated in \autoref{tab:example}.
This formulation mitigates the limitations of independent pairwise matching and fits real-world ER scenarios.
First, current SOTA blocking methods adhere to the k-nearest neighbor (kNN) search paradigm, which retrieves a list of potential matches for each entity record.
In addition, this formulation accommodates both single-source deduplication and dual-source record linkage, and makes good use of the one-to-one assumption, \ie record $r$ matches at most one of the records in potential matches $R$.
This assumption is widespread in record linkage, and deduplication with canonicalization.

\subsection{LLM as a Matcher}
\label{sec:matching}

Recent work formulates entity matching as a binary classification task based on LLMs~\citep{DBLP:journals/pvldb/NarayanCOR22,DBLP:conf/adbis/PeetersB23,DBLP:conf/icde/FanHFC00024,DBLP:journals/corr/abs-2401-03426,DBLP:conf/edbt/PeetersSB25}.
In this strategy, an LLM acts as a pairwise matcher to determine whether two records match.
Specifically, given an entity record $r$ and its potential matches $R = \{r_1, r_2, \ldots, r_n \}$, this approach independently classifies each pair of records $(r, r_i)_{1 \le i \le n}$ as matching or not by interfacing LLMs with an appropriate matching prompt, as shown in \autoref{fig:introduction}(a):
\begin{equation*}
  \operatorname{LLM}_m \colon \left\{(r, r_i) \mid r_i \in R \right\} \rightarrow \left\{\mathrm{Yes}, \mathrm{No}\right\}
\end{equation*}

Unlike previous studies, the core of LLM-based applications is to prompt LLMs to generate the correct answer, namely prompt engineering.
An appropriate prompt should include the task instruction, such as ``\textit{Do these two records refer to the same real-world entity? Answer Yes or No}''.
Optionally, a prompt could include detailed rules or several in-context learning examples to guide LLMs in performing this task.
Given the need for long contexts in other strategies, and the instability of existing prompt engineering methods for entity matching~\cite{DBLP:conf/edbt/PeetersSB25}, we only attempt few-shot prompting for the matching strategy and leave the exploration of better prompt engineering with different strategies to future work.

This independent matching strategy ignores the global consistency of ER, as well as the capabilities of LLMs to incorporate record interactions.
The traditional solution to satisfy these constraints is to construct a graph based on the similarity scores $s_i$ of record pairs $(r, r_i)$ and to further cluster on the similarity graph.
We can obtain the similarity scores from LLMs by calibrating the generated probabilities $p$ of labels~\cite{DBLP:journals/tmlr/LiangBLTSYZNWKN23}.
Formally, the similarity score $s_i$ can be defined as:
\begin{equation*}
  s_{i}=
  \begin{cases}
    1 + p(\mathrm{Yes} \mid (r, r_i)), & \text{if generate ``Yes''} \\
    1 - p(\mathrm{No} \mid (r, r_i)), & \text{if generate ``No''}
  \end{cases}
\end{equation*}
Unfortunately, the generation probabilities are not available for many black-box commercial LLMs.
Moreover, the probabilities on short-form labels are misaligned for common open-source chat-tuned LLMs because they are fine-tuned to respond in detail.
The need to investigate better strategies for LLM-based entity matching arises in ER.

\subsection{LLM as a Comparator}
\label{sec:comparing}

In this section, we introduce a comparing strategy for LLM-based entity matching that simultaneously compares two potential matches to a given record.
Specifically, given an entity record $r$ and its potential matches $R = \{r_1, r_2, \ldots, r_n \}$, the comparing strategy compares two records $r_i$ and $r_j$ from potential matches $R$ to determine which is more consistent with record $r$ by interfacing LLMs with a comparison prompt, as shown in \autoref{fig:introduction}(b):
\begin{equation*}
  \operatorname{LLM}_c \colon \left\{(r, r_i, r_j) \mid r_{i,j} \in R \right\} \rightarrow \left\{\mathrm{A}, \mathrm{B}\right\}
\end{equation*}
where $\mathrm{A}$ and $\mathrm{B}$ are labels corresponding to record $r_i$ and $r_j$.
Since LLMs may be sensitive to the prompt order, we compare the record pair $(r_i, r_j)$ to record $r$ twice by swapping their order.

Compared to the matching strategy, the comparing strategy introduces an additional record for more record interactions and shifts the task paradigm.
It focuses on indicating the relative relationship between two potential matches of a given record, rather than making a direct match or no match decision.
Therefore, this strategy is suitable for ranking and fine-grained filtering to determine the most likely records for identification.

To rank candidate records using the comparing strategy, we can compute similarity scores to estimate how closely each candidate matches the anchor record.
Unlike the matching strategy, the comparing strategy can obtain similarity scores of record pairs using black-box LLMs, which do not provide probabilities.
In such case, the similarity score $s_i$ of record pair $(r, r_i)$ can be defined as:
\begin{equation*}
  s_i = 2 \times \sum_{j \neq i} \mathbb{1}_{r_i >_{r} r_j} + \sum_{j \neq i} \mathbb{1}_{r_i =_{r} r_j}
\end{equation*}
where $\mathbb{1}_{r_i >_{r} r_j}$ and $\mathbb{1}_{r_i =_{r} r_j}$ indicate that record $r_i$ wins twice and once in comparison with record $r_j$ to record $r$.
When LLMs do provide probabilities, the similarity score $s_i$ can be defined as:
\begin{equation*}
    s_i = \sum_{j \ne i} \left(p(\mathrm{A}\mid(r, r_i, r_j)) + p(\mathrm{B}\mid(r, r_j, r_i)) \right)
\end{equation*}

However, the advantage of the comparing strategy in obtaining similarity scores comes at the cost of using LLMs as the basic unit of comparison and $\mathcal{O}(n^2)$ complexity.
Fortunately, for entity matching, we only care about a small number of most similar candidates, and there are many comparison sort algorithms available to find the top-$k$ elements efficiently.
In this paper, we use the \emph{bubble sort} algorithm to find the top-$k$ elements, optimizing the complexity of the comparing strategy to $\mathcal{O}(kn)$.
To avoid confusion, we refer to the comparison of all pairs as \texttt{comparing\textsubscript{all-pair}} in our experiments.

\begin{table}
  \centering
  \renewcommand{\arraystretch}{1.1}
  \setlength{\tabcolsep}{0.3\tabcolsep}
  \resizebox{0.98\columnwidth}{!}{%
    \begin{tabular}{@{}ccccc@{}}
      \toprule
      \textbf{Strategy}
      & \textbf{\begin{tabular}[c]{@{}c@{}} Similarity \\ Score \end{tabular}}
      & \textbf{\begin{tabular}[c]{@{}c@{}} Interaction \\ Level \end{tabular}}
      & \textbf{\begin{tabular}[c]{@{}c@{}} \# LLM \\ Invocations \end{tabular}}
      & \textbf{\begin{tabular}[c]{@{}c@{}} \# Input \\ Records \end{tabular}} \\ \midrule
      Matching  & \bminus    & \bplus             & $\mathcal{O}(n)$   & $2n$            \\
      Comparing & \ding{51}  & \bplus\bplus       & $\mathcal{O}(kn)$  & $3k(2n-k-1)$    \\
      Selecting & \ding{55}  & \bplus\bplus\bplus & $\mathcal{O}(1)$   & $n+1$         \\ \bottomrule
    \end{tabular}%
  }
  \caption{
    Comparison of different strategies.
    ``\textbf{--}'' means that the matching strategy can only calibrate similarity scores if the generation probability is available.
     ``\# LLM Invocations'' and ``\# Input Records'' represent the number of (\#) LLM invocations and records input to LLMs using different strategies for record $r$ and its $n$ potential matches $R$, respectively.
    $k$ denotes the number of top candidates considered by the comparing strategy.
  }
  \label{tab:strategies}
\end{table}

\subsection{LLM as a Selector}
\label{sec:selecting}

In this section, we introduce a selecting strategy that uses an LLM to select the matching record of a given record from a list of potential matches.
Specifically, given an entity record $r$ and its potential matches $R = \{r_1, r_2, \ldots, r_n \}$, this strategy directly selects the match of record $r$ from $R$ by interfacing LLMs with an appropriate selection prompt, as shown in \autoref{fig:introduction}(c):
\begin{equation*}
  \operatorname{LLM}_s \colon \left\{(r, R) \right\} \rightarrow \left\{1, 2, \ldots, n\right\}
\end{equation*}
where $1, \ldots, n$ indicates the corresponding record.

In this way, LLMs can be explicitly required to identify only one match per record $r$ from the potential matches $R$.
Furthermore, feeding LLMs all potential matches in the same context at a time allows LLMs to make better decisions by considering interactions between records.
Using \autoref{tab:example} as an example, it is easier for LLMs to recognize the less critical attributes, such as authors, and identify the third record as the true match by comparing the values of title and year across different records.

One challenge in applying the selecting strategy to LLM-based entity matching is that sometimes there is no actual match of record $r$ in potential matches $R$.
A trivial solution to this challenge could be to perform a pairwise matching after the selection, which would undermine the advantages of the selecting strategy.
Another method could be to add ``none of the above'' as an additional option to allow LLMs to refuse to select any record from the potential matches, which can be formulated as:
\begin{equation*}
  \operatorname{LLM}_{s_N} \colon \left\{(r, R) \right\} \rightarrow \left\{0, 1, 2, \ldots, n\right\}
\end{equation*}
where $0$ indicates the ``none of the above'' option.

However, the selecting strategy relies heavily on the capabilities of LLMs for fine-grained understanding and implicit ranking in long contexts.
Our experimental results show that current LLMs suffer from position bias, with the selection accuracy varying significantly as the position of the matching record increases in the candidate list (\autoref{sec:ex-strategy}).
In practice, the recall-oriented blocking step often generates a considerable number of potential matches for each record, exceeding the context length that LLMs can effectively reason~\citep{DBLP:conf/acl/LevyJG24}.
Therefore, it is a challenge to mitigate the position bias and the long context requirement for the selecting strategy.

\subsection{Compound Entity Matching Framework}
\label{sec:comem}

\begin{figure}
  \centering
  \includegraphics[width=\columnwidth]{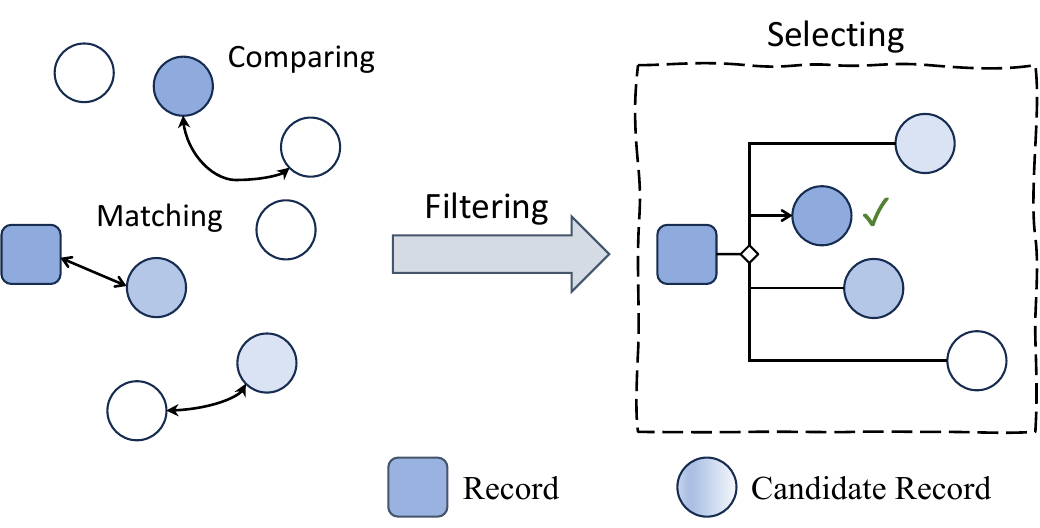}
  \caption{
    Illustration of \sys.
    It first filters candidate records by matching or comparing strategies and then identifies the match via the selecting strategy.
  }
  \label{fig:hybrid}
\end{figure}

Based on the advantages and shortcomings of different strategies, we further propose a compound entity matching framework (\sys).
\sys addresses various challenges in LLM-based entity matching by integrating the advantages of different strategies and LLMs.
\autoref{tab:strategies} shows a comparison of these strategies.
The matching and comparing strategies are applicable for local ranking, while the selecting strategy is suitable for fine-grained identification.
Therefore, as shown in \autoref{fig:hybrid}, we first utilize a medium-sized LLM to rank and filter potential matches $R$ of record $r$ with the matching or comparing strategy.
We then utilize an LLM to identify the match of record $r$ from only the top $k$ candidates with the selecting strategy.

Our \sys integrates the advantages of different strategies through a filtering then identifying pipeline.
It first utilizes the local matching or comparing strategy to rank potential matches for preliminary screening, which can effectively mitigate the position bias and the long context requirement of the selecting strategy.
It then utilizes the global selecting strategy to incorporate record interactions for fine-grained optimization, which can effectively mitigate the consistency ignorance of the matching strategy.
Therefore, \sys is able to strike a balance between entity matching requirements and current LLM capabilities, achieving significant performance improvements.

By integrating LLMs of different sizes, \sys can also effectively reduce the cost of LLM invocations for entity matching.
In practice, direct use of commercial LLMs is expensive because entity matching is a computationally intensive task.
\sys delegates a significant part of the computation to medium-sized LLMs.
Experimental results show that the ranking process can be performed well by using open-source medium-sized (3B\textasciitilde11B) LLMs (\autoref{sec:ex-llm}).
As a result, the proper integration of LLMs not only improves the performance of entity matching, but also reduces the cost for practical application.


\begin{table}
  \centering
  \setlength{\tabcolsep}{0.6\tabcolsep}
  \resizebox{0.98\columnwidth}{!}{%
    \begin{tabular}{@{}ccrrrr@{}}
      \toprule
      \textbf{Dataset} & \textbf{Domain} & \textbf{\# D1} & \textbf{\# D2} & \textbf{\# Attr} & \textbf{\# Pos} \\ \midrule
      Abt-Buy (AB)          & Product         & 1076           & 1076           & 3                & 1076            \\
      Amazon-Google (AG)    & Software        & 1354           & 3039           & 4                & 1103            \\
      DBLP-ACM (DA)         & Citation        & 2616           & 2294           & 4                & 2224            \\
      DBLP-Scholar (DS)     & Citation        & 2516           & 61353          & 4                & 2308            \\
      IMDB-TMDB (IM)        & Movie           & 5118           & 6056           & 5                & 1968            \\
      IMDB-TVDB (IV)        & Movie           & 5118           & 7810           & 4                & 1072            \\
      TMDB-TVDB (TT)        & Movie           & 6056           & 7810           & 6                & 1095            \\
      Walmart-Amazon (WA)   & Electronics     & 2554           & 22074          & 6                & 853             \\ \bottomrule
    \end{tabular}%
  }
  \caption{Statistics of experimental datasets. \# denotes ``number of'', D1 and D2 represent records from the 1st and 2nd sources, respectively. Attr and Pos refer to \emph{attributes} of structured records and \emph{positive} (matching) record pairs, respectively.}
  \label{tab:statistics}
\end{table}

\section{Experiments}
\label{sec:exp}

In this section, we conduct thorough experiments to investigate three strategies for LLM-based entity matching.
First, we present the main experimental results (\autoref{sec:ex-main}).
Next, we perform the analysis of different strategies (\autoref{sec:ex-strategy}).
Finally, we examine the effect of different LLMs on these strategies (\autoref{sec:ex-llm}).

\begin{table*}
  \centering
  \setlength{\tabcolsep}{0.7\tabcolsep}
  \resizebox{0.95\textwidth}{!}{%
    \begin{tabular}{l@{\hspace{4\tabcolsep}}cccccccc@{\hspace{4\tabcolsep}}c@{\hspace{2\tabcolsep}}c}
      \toprule
      \multicolumn{1}{l}{\textbf{}} & \textbf{AB}    & \textbf{AG}    & \textbf{DA}    & \textbf{DS}    & \textbf{IM}    & \textbf{IV}    & \textbf{TT}    & \textbf{WA}    & \textbf{Mean}  & \textbf{Cost} \\ \midrule
      \multicolumn{11}{l}{\textit{Supervised}}                                                                                                                                                 \\
      Ditto~\cite{DBLP:journals/pvldb/0001LSDT20}                & 77.34          & 63.79          & 93.80          & 90.02          & 97.06          & 78.59          & 87.15          & 57.75          & 80.69          & 0.29          \\
      HierGAT~\cite{DBLP:conf/sigmod/0002GC0L22}              & 75.51          & 64.45          & 98.01          & 89.47          & 97.69          & 77.73          & 85.34          & 78.55          & 83.34          & 1.10          \\ \cmidrule{1-11}\morecmidrules\cmidrule{1-11}
      \multicolumn{11}{l}{\textit{Un/Self-supervised}}                                                                                                                                                               \\
      ZeroER~\cite{DBLP:conf/sigmod/WuCSCT20}               & 32.66          & 22.14          & \textbf{99.32} & 84.14          & 43.32          & 0.50           & 53.76          & 61.52          & 49.67          & /             \\
      Sudowoodo~\cite{DBLP:conf/icde/Wang0023}            & 58.82          & 50.45          & 90.97          & 77.06          & 84.72          & 71.88          & 76.32          & 52.36          & 70.32          & 0.63          \\ \cmidrule{1-11}
      \multicolumn{11}{l}{\textit{GPT-3.5 Turbo}}                                                                                                                                         \\
      Matching~\cite{DBLP:conf/edbt/PeetersSB25}             & 56.03          & 44.36          & 78.93          & 71.89          & 72.05          & 61.11          & 77.05          & 50.77          & 64.02          & 4.52          \\
      Matching {\small(6-shot)}~\cite{DBLP:conf/edbt/PeetersSB25}            & 77.59          & 60.21          & 73.13          & 52.88          & 84.05          & 71.45          & 71.21          & 69.37          & 69.99          & 32.75         \\
      Comparing            & 79.45          & 51.61          & 76.61          & 65.59          & 62.92          & 46.12          & 87.27          & 65.34          & 66.86          & 11.75         \\
      Selecting            & 80.31          & 63.65          & 88.62          & 80.61          & 92.43          & 83.36          & 83.66          & 80.18          & 81.60          & 1.71          \\
      \sys                 & 87.62          & {\ul 69.63}    & 90.85          & 84.68          & \textbf{96.74} & \textbf{84.16} & 84.82          & {\ul 86.37}    & {\ul 85.61}    & 0.92          \\ \cmidrule(lr){1-11}
      \multicolumn{11}{l}{\textit{GPT-4o Mini}}                                                                                                                                           \\
      Matching~\cite{DBLP:conf/edbt/PeetersSB25}             & 81.37          & 51.95          & 61.28          & 48.76          & 89.64          & 61.65          & 72.84          & 74.92          & 67.80          & 0.46          \\
      Matching {\small(6-shot)}~\cite{DBLP:conf/edbt/PeetersSB25}             & 83.03          & 63.63          & 84.50          & 71.37          & {\ul 95.70}    & 71.82          & 72.54          & 80.97          & 77.94          & 3.21          \\
      Comparing            & \textbf{89.24} & 65.61          & {\ul 93.04}    & \textbf{88.41} & 89.86          & 75.05          & {\ul 89.80}    & 83.85          & 84.36          & 1.21          \\
      Selecting            & 82.37          & 69.01          & 83.28          & 81.11          & 94.43          & {\ul 83.61}    & 87.58          & 76.66          & 82.26          & 0.17          \\
      \sys                 & {\ul 88.24}    & \textbf{71.47} & 90.58          & {\ul 87.84}    & 95.62          & 78.07          & \textbf{90.97} & \textbf{88.56} & \textbf{86.42} & 0.09          \\ \bottomrule
    \end{tabular}%
  }
  \caption{
    Overall performance and cost of different methods. We bold the \textbf{best} F1 score and underline the {\ul{second best}} for non-supervised methods. The cost of learning-based methods includes both the training and testing GPU costs.
  }
  \label{tab:main-result}
\end{table*}


\subsection{Experimental Setup}

\stitle{Datasets.}
We focused on record linkage, a common form of entity resolution that identifies matching records between two data sources.
Specifically, we used eight clean-clean ER datasets collected by pyJedAI~\cite{DBLP:conf/semweb/Nikoletos0K22}.
\autoref{tab:statistics} shows the statistics of these datasets, where record collections \texttt{D1} and \texttt{D2} represent records from the first and second sources, respectively.
For each dataset, we applied the SOTA blocking method Sparkly~\cite{DBLP:journals/pvldb/PaulsenGD23} as preprocessing to retrieve 10 potential matches from \texttt{D2} for each record in \texttt{D1}.
The recall@10 of Sparkly on all datasets ranges from 86.57\% to 99.96\%, demonstrating its effectiveness in retrieving potential matches.
We sampled 400 records from \texttt{D1} for evaluation, 300 of which had matches, and formed 4,000 pairs of records by combining them with their potential matches from \texttt{D2}.
To build training sets for learning-based methods, we further sampled 5,000 record pairs from the remaining records and their potential matches.
Through this process, we constructed entity matching datasets that satisfied our formulation, with all methods evaluating on the same datasets after blocking to ensure a fair comparison.

\stitle{Baseline.}
We considered several SOTA methods as our baselines, including the unsupervised ZeroER~\cite{DBLP:conf/sigmod/WuCSCT20}, the self-supervised Sudowoodo~\cite{DBLP:conf/icde/Wang0023}, and the LLM-based matching strategy~\cite{DBLP:conf/edbt/PeetersSB25}.
For a comprehensive comparison, we also included two representative supervised methods, Ditto~\cite{DBLP:journals/pvldb/0001LSDT20} and HierGAT~\cite{DBLP:conf/sigmod/0002GC0L22}.\footnote{We followed their open-source implementations and default parameters for reproduction.}

\stitle{Evaluation Metrics.}
Consistent with prior studies, we report the F1 score as performance measure.
We also report the cost (\$) of LLM invocations.
For the compute cost of open-source LMs, we estimated it based on the training or inference time required and the \href{https://www.runpod.io/gpu-instance/pricing}{hourly price} of the cloud NVIDIA A40.

\stitle{Implementation Details.}
We used GPT-4o Mini {\small\textsf{(0718)}} and GPT-3.5 Turbo {\small\textsf{(0613)}} as the main LLMs for analysis.
We also examined the effect of eight open-source \emph{instruction-tuned} LLMs, including Llama-3.1-8B~\citep{DBLP:journals/corr/abs-2407-21783}, Qwen2-7B~\citep{DBLP:journals/corr/abs-2407-10671}, Mistral-7B~\citep{DBLP:journals/corr/abs-2310-06825}, Mixtral-8x7B~\citep{DBLP:journals/corr/abs-2401-04088}, \href{https://huggingface.co/CohereForAI/c4ai-command-r-v01}{Command-R-35B}, Flan-T5-XXL~\citep{DBLP:journals/jmlr/ChungHLZTFL00BW24}, Flan-UL2~\citep{DBLP:conf/iclr/Tay00GW0CBSZZHM23} and Solar-10.7B~\citep{DBLP:conf/naacl/KimKPLSKKKLKAYLPGCLK24}.
The specific prompts can be found in \autoref{app:prompt}, with the generation temperature of all LLMs set to 0 for reproducibility.
For in-context learning, we retrieve 3 positives and 3 negatives as few-shot examples based on record similarity as \citet{DBLP:conf/edbt/PeetersSB25}.
Since the comparing strategy produces only relative orders, we applied the matching strategy to the top 1 candidate after bubble sort comparing.
In \sys, we used Flan-T5-XL to rank candidates with the matching strategy and kept the top 4 candidates for selection.

\subsection{Main Results}
\label{sec:ex-main}

We first compare the performance and cost of different methods, with the following findings.

\begin{figure*}
  \centering
  \includegraphics[width=0.95\textwidth]{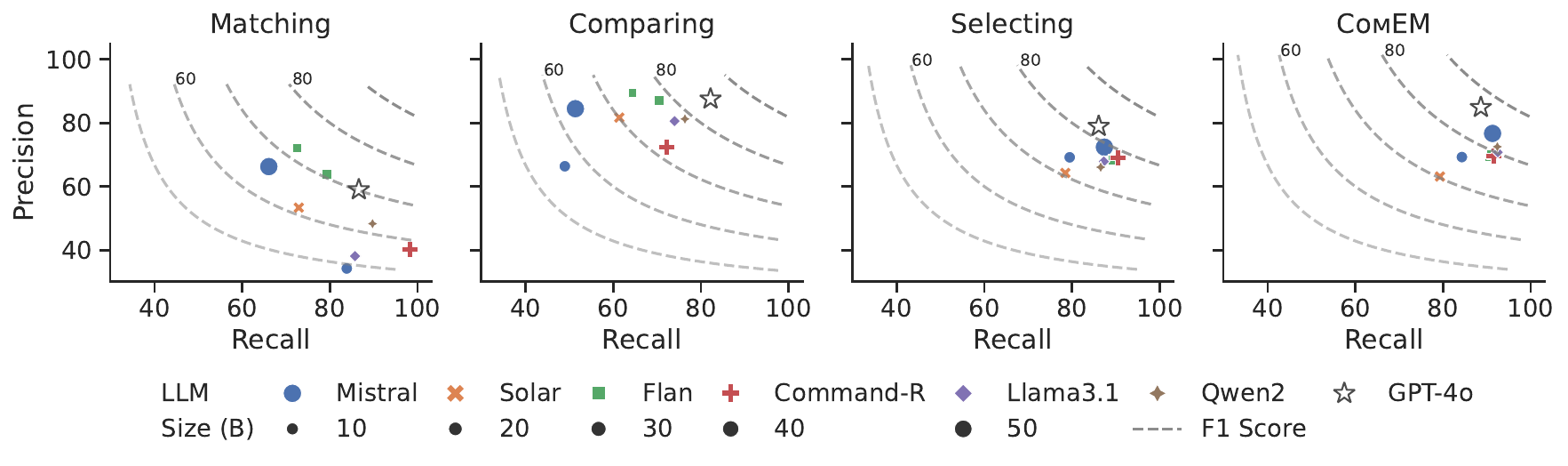}
  \caption{Effect of open-source LLMs on different strategies and \sys.}
  \label{fig:llms}
\end{figure*}

\textbf{Finding 1.} \emph{Incorporating record interactions is essential for LLM-based entity matching.}
As shown in \autoref{tab:main-result}, the performance of LLM-based entity matching increases with incorporating record interaction.
The comparing strategy outperforms the independent matching strategy by an average of 10.7\% F1 score, and the selecting strategy further improves the F1 score by an average of 5.32\% over the comparing strategy.
The advantages of the comparing and selecting strategies over the matching strategy are also evident across different LLMs in \autoref{fig:llms}.
To further verify that these improvements are due to the strategy, we perform 6-shot matching, ensuring that the number of records is consistent with the selecting strategy.
We can see that the selecting strategy still outperforms 6-shot matching by 7.97\% in F1.
Moreover, the proposed strategies enable LLM-based entity matching to surpass SOTA un/self-supervised methods and to be comparable to supervised methods that require extensive labeling data.
\emph{These results highlight the effectiveness of our proposed strategies and open new avenues for LLM-based entity matching.}

\textbf{Finding 2.} \emph{By integrating the advantages of different strategies and LLMs, \sys can accomplish entity matching more effectively and cost-efficiently.}
As shown in \autoref{tab:main-result}, compared to the single comparing and selecting strategies, \sys achieves 2\textasciitilde18\% F1 improvements while spending less.
The filtering and identifying pipeline improves precision considerably without sacrificing the high recall of the selecting strategy.
These results reveal that integrating multiple strategies can complement single strategies and mitigate the position bias of the selecting strategy in long contexts.
However, using a single powerful but costly commercial LLM to complete the entire pipeline obscures the cost efficiency of the selecting strategy.
By introducing a medium-sized LLM for preliminary filtering, \sys improves performance while spending less than direct selection.
\emph{As a result, \sys underscores the importance of task decomposition and LLM composition, illuminating an effective route for compound EM using LLMs.}

\begin{figure}
  \includegraphics[width=\columnwidth]{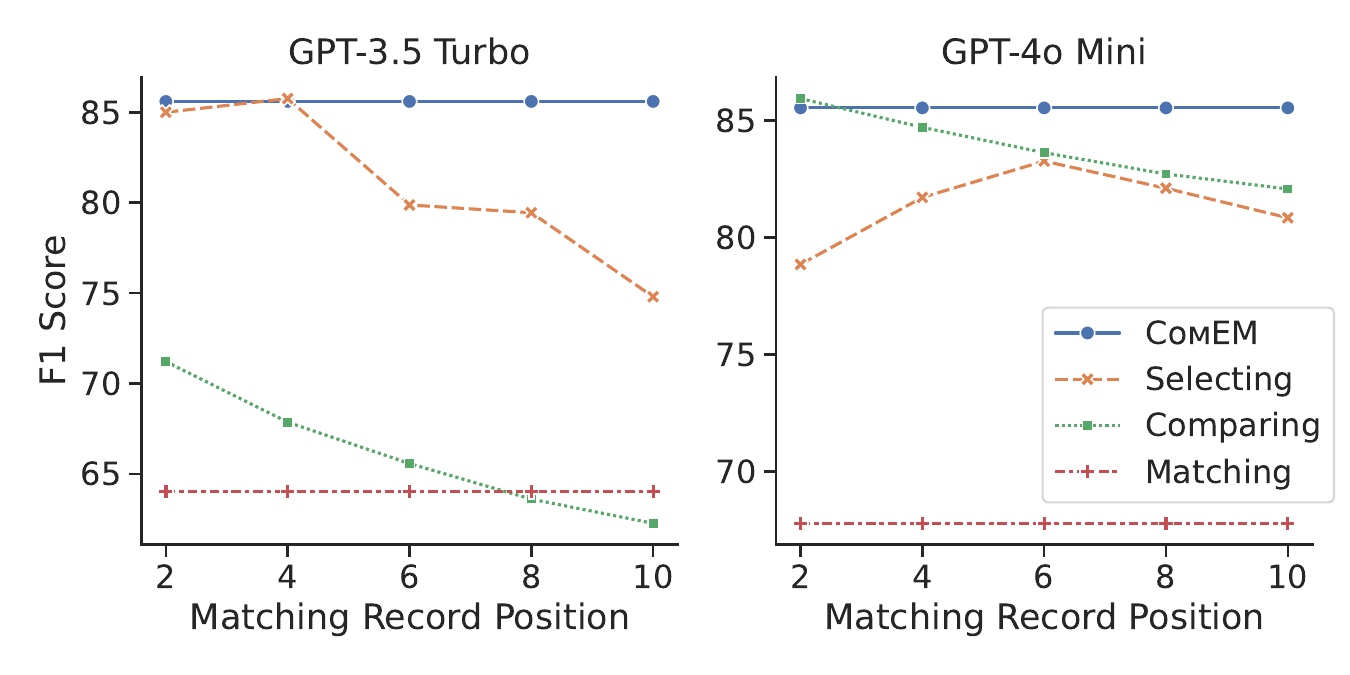}
  \caption{F1 score \wrt matching record positions.}
  \label{fig:position}
\end{figure}

\subsection{Analysis of Strategies}
\label{sec:ex-strategy}

We then analyze the advantages and shortcomings of different strategies from different perspectives.

\textbf{Finding 3.} \emph{The selecting strategy is the most cost-effective strategy for LLM-based entity matching.}
Monetary cost is also an important factor when interfacing LLMs for entity matching in practice, as it is computationally intensive.
As shown in \autoref{tab:main-result}, the selecting strategy costs less than half of the matching strategy.
This is because the selecting strategy saves $n-1$ times of repeatedly inputting anchor records and task instructions into LLMs.
The comparing strategy, however, considers two potential matches at a time and interfaces the LLM twice, making its cost more than twice that of the matching strategy.
Therefore, the selecting strategy stands out for its effectiveness and efficiency.

\textbf{Finding 4.} \emph{Strategies that incorporate multiple records suffer from the position bias of LLMs.}
As shown in \autoref{fig:position}, the performance of the comparing and selecting strategies varies significantly as the position of the matching records moves down in the candidate list.
For the comparing strategy optimized with bubble sort, matching records cannot be ranked at the top if there is any incorrect comparison.
The selecting strategy is also highly sensitive to the matching record positions, while \sys can alleviate this.
Therefore, the position bias of LLMs limits the performance and generality of the comparing and selecting strategies.

\subsection{Effect of LLMs}
\label{sec:ex-llm}

We further examine the effect of open-source LLMs on these strategies to identify matches or rank.

\textbf{Finding 5.} \emph{There is no single LLM that is uniformly dominant across all strategies.}
\autoref{fig:llms} shows the efficacy of proposed strategies for open-source LLMs, with detailed results in \autoref{app:llm}.
We can see that the F1 scores of the matching, comparing, and selecting strategies for different LLMs mostly fall between 50\%\textasciitilde70\%, 60\%\textasciitilde80\%, and 70\%\textasciitilde80\%, respectively.
In general, similar to GPT-3.5 Turbo, the comparing strategy is better than the matching strategy, while the selecting strategy is further better than the comparing strategy.
The consistent performance between strategies confirms the effectiveness of incorporating record interactions in these ways.
Concretely, some chat LLMs, such as Llama3-8B and Mistral-7B, produce numerous false positives and thus perform poorly with the matching strategy.
Nevertheless, they achieve significant improvement and satisfactory performance by comparing and selecting strategies, respectively.
Moreover, although Flan-T5-XXL and Flan-UL2 lag behind GPT-4o by about 5\% F1 in the selecting strategy, we find that they perform quite well in the matching and comparing strategies.
These task-tuned LLMs follow instructions better than chat-tuned LLMs, and can output only the requested labels instead of long-form responses, making it convenient to utilize label generation probabilities.
In conclusion, there is a noticeable variance in the capabilities of different LLMs for a single strategy, and the efficacy of different strategies for a single LLM can also be significantly distinct.

\begin{figure}
  \centering
  \includegraphics[width=0.9\columnwidth]{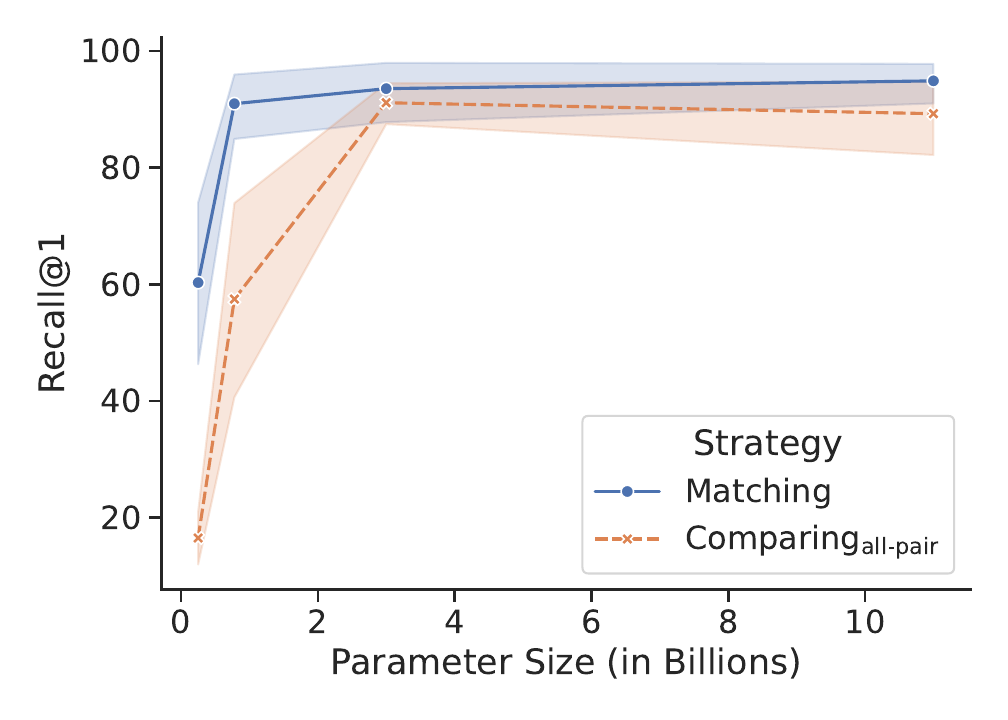}
  \caption{Ranking recall@1 \wrt model parameters.}
  \label{fig:matching_comparing}
\end{figure}

\textbf{Finding 6.} \emph{Matching strategy is better for ranking and filtering than comparing strategy.}
The superiority of Flan-T5 in the matching and comparing strategies leads us to explore the possibility of using it to rank and filter potential matches for the selecting strategy.
As shown in \autoref{fig:matching_comparing}, the matching strategy outperforms the comparing strategy under different model parameter sizes, even though the latter performs $\mathcal{O}(n^2)$ comparisons.
The difference is small on Flan-T5-XL (3B) and Flan-T5-XXL (11B), but significant on smaller models.
This may be due to the fact that these models are trained on many pairwise tasks, such as natural language inference and question answering, but few triplewise tasks.
Therefore, in terms of effectiveness and efficiency, the matching strategy is more suitable for ranking and filtering potential matches.

\begin{figure}
  \includegraphics[width=\columnwidth]{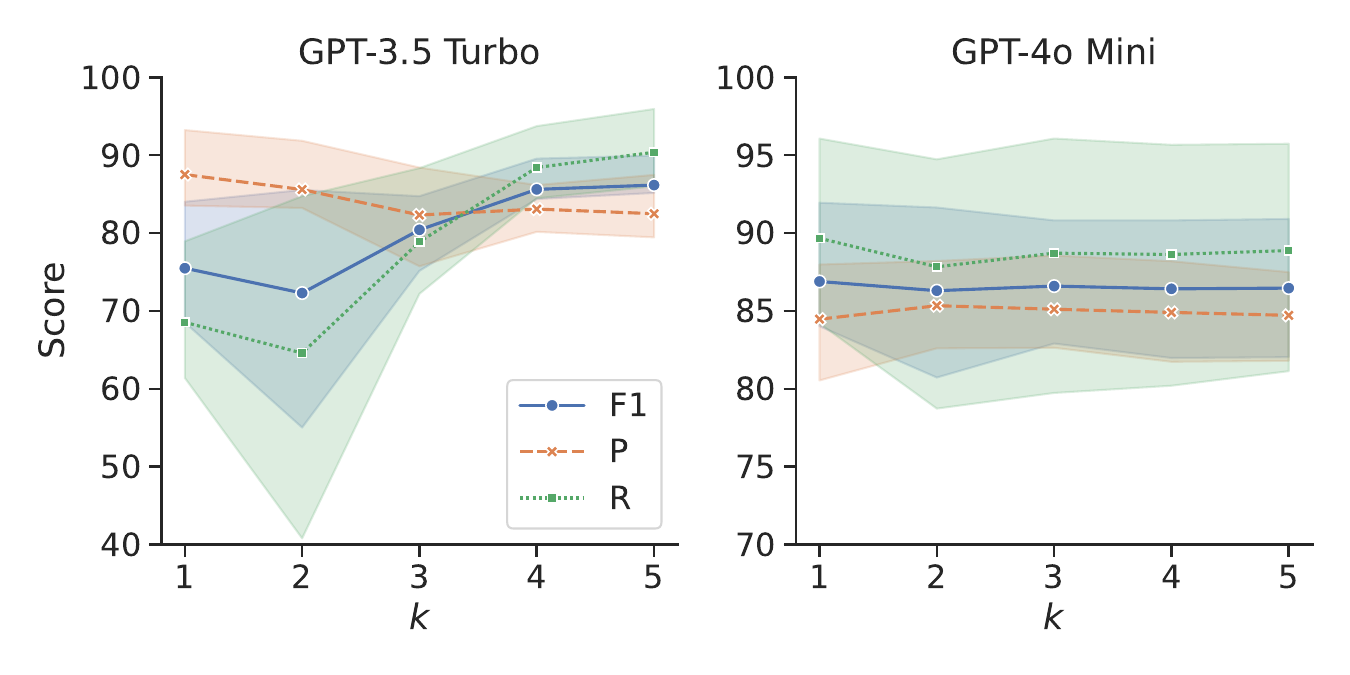}
  \caption{Average F1, Precision, and Recall \wrt number of candidate retained ($k$) for further selection.}
  \label{fig:top_k}
\end{figure}

\subsection{Ablation Study}
We perform ablation studies on the number of candidate records for further identification.
As shown in \autoref{fig:top_k}, the performance of GPT-3.5 Turbo is highly variable, while that of GPT-4o Mini is relatively stable.
These results suggest that as LLMs evolve, \sys may become more robust to the number of potential matches retained.


\section{Conclusion}

In this paper, we investigate three strategies of LLMs for entity matching to bridge the gap between local matching and global consistency of ER.
Our research shows that incorporating record interactions is essential for LLM-based entity matching.
By examining the effect of broad LLMs on these strategies, we further design a \sys framework that integrates the advantages of multiple strategies and LLMs.
The effectiveness and cost-efficiency of \sys highlight the importance of task decomposition and LLM composition, opening up new avenues for entity matching using LLMs.


\section*{Limitations}

This study aims to investigate different strategies for LLM-based entity matching.
We conducted thorough experiments with two commercial LLMs and eight open-source LLMs to provide a broad base for our analysis.
The selection of LLMs is based on considerations of popularity, availability, and cost.
Future research could explore whether similar findings hold as LLMs evolve and how performance changes relative to our results.

Since LLMs were trained on massive amount of web data, they are likely to have seen similar and same records, or even some matching results, even though the labels of the matches are stored separately.
Nevertheless, the performance of these strategies is relatively consistent across 10 LLMs and varies greatly for the same LLM when using different strategies, highlighting that data exposure is not the determining factor in their effectiveness.
In the future, it will be valuable to evaluate LLM-based entity matching on new or non-public data.

The investigation of different strategies was conducted using basic zero- or few-shot prompting, a simple and effective paradigm for applying LLMs.
We could not ignore the role of potential advanced prompt engineering methods in improving the accuracy and robustness of LLMs.
In addition, fine-tuning LLMs for better execution of different strategies is also a worthwhile direction.

Finally, we have demonstrated the effectiveness of the compound framework in entity matching that integrates different strategies and LLMs.
We would like to continue to develop specific modules for entity matching and extend this paradigm to different stages of entity resolution.


\section*{Acknowledgments}
We sincerely thank the reviewers for their insightful comments and valuable suggestions. This work was supported by the Natural Science Foundation of China (No. 62122077, 62106251) and Beijing Natural Science Foundation (L243006).

\bibliography{src/ref}
\clearpage

\appendix

\begin{table}
  \centering
  \setlength\extrarowheight{-3pt}
  \setlength{\defaultaddspace}{3pt}
  \resizebox{\columnwidth}{!}{%
    \small
    \begin{tabular}{@{}c@{}}
      \toprule
      \textbf{\textit{Matching}} \\ \addlinespace
      \begin{tabular}[c]{@{}p{\columnwidth}@{}}Do the two entity records refer to the same real-world entity? Answer "Yes" if they do and "No" if they do not.\\ \\ Record 1: \{\{ record\_left \}\}\\ Record 2: \{\{ record\_right \}\}\end{tabular}
      \\ \midrule
      \textbf{\textit{Comparing}} \\ \addlinespace
      \begin{tabular}[c]{@{}p{\columnwidth}@{}}Which of the following two records is more likely to refer to the same real-world entity as the given record? Answer with the corresponding record identifier "Record A" or "Record B".\\ \\ Given entity record: \{\{ anchor \}\}\\ \\ Record A: \{\{ candidate\_left \}\}\\ Record B: \{\{ candidate\_right \}\}\end{tabular}
      \\ \midrule
      \textbf{\textit{Selecting}} \\ \addlinespace
      \begin{tabular}[c]{@{}p{\columnwidth}@{}}Select a record from the following candidates that refers to the same real-world entity as the given record. Answer with the corresponding record number surrounded by "{[}{]}" or "{[}0{]}" if there is none.\\ \\ Given entity record: \{\{ anchor \}\}\\ \\ Candidate records:\{\% for candidate in candidates \%\}\\ {[}\{\{ loop.index \}\}{]} \{\{ candidate \}\}\{\% endfor \%\}\end{tabular} \\ \bottomrule
    \end{tabular}%
  }
  \caption{
    Specific prompts of different strategies.
    We use JinJa template syntax to display the placeholders for the \emph{anchor} record and potential matches (\emph{candidates}).
  }
  \label{tab:prompts}
\end{table}

\section{Global Consistency of ER}
\label{app:global_consistency}

In this paper, we refer to the interdependence of matching decisions in entity resolution as global consistency.
This means that whether two records match is not an isolated decision, but is influenced by the matching results of other record pairs.
This includes properties such as: 1) \textit{Reflexive}: A record always matches itself; 2) \textit{Symmetric}: If record A matches record B, then B also matches A; 3) \textit{Transitive}: If A matches B, and B matches C, then A should match C; 4) \textit{Mutually exclusive}: In some cases such as clean-clean ER, if A matches B, it cannot match C.
Global consistency motivates the incorporation of more record interactions for LLM-based entity matching, rather than just considering two records independently.

\section{Strategy Prompts}
\label{app:prompt}

The prompts for different strategies of LLM-based entity matching used in this paper are presented in \autoref{tab:prompts}.
To ensure fairness, the same prompts were used for all experimental LLMs.
These prompts were constructed through a manual process of prompt engineering, which involved the testing and comparing of multiple variations to determine the most effective ones.
In addition to the task description, we included specific response instructions such as ``\emph{Answer "Yes" if they do and "No" if they do not}'' to guide the responses of LLMs.
For in-context learning, prompts and labels were repeatedly inputted for each example, followed by the records to be matched.
We post-processed the LLM responses to obtain the final predicted labels.

\section{Detailed Results of Open-Source LLMs under Different Strategies}
\label{app:llm}

We provide the detailed F1 scores of open-source LLMs under different strategies and \sys in \autoref{tab:open-llm}.
Among the eight LLMs evaluated in our experiment, six achieve the best performance through the selecting strategy, and two achieve better performance through the comparing strategy.
In summary, our proposed strategies are universally applicable across different LLMs for entity matching.
We have observed that it is difficult to limit the output of many chat-tuned LLMs simply by prompts, which may affect their actual performance.
Therefore, how to calibrate the label probabilities from the long-form responses of LLMs is also important for performance improvement.

\begin{table*}
  \centering
  \aboverulesep=0ex
  \belowrulesep=0ex
  \resizebox{\textwidth}{!}{%
    \begin{tabular}{c|c|cccccccc|c}
      \hline
      \textbf{LLM}                           & \textbf{Strategy} & \textbf{AB} & \textbf{AG} & \textbf{DA} & \textbf{DS} & \textbf{IM} & \textbf{IV} & \textbf{TT} & \textbf{WA} & \textbf{Mean} \\ \hline
      \multirow{4}{*}{Mistral-Instruct-7B}   & Matching          & 40.70       & 37.77       & 24.68       & 28.89       & 64.86       & 64.49       & 49.91       & 55.96       & 45.91         \\
                                             & Comparing         & 54.68       & 32.10       & 49.28       & 49.75       & 74.38       & 52.25       & 81.69       & 44.39       & 54.82         \\
                                             & Selecting         & 67.26       & 57.31       & 83.36       & 74.27       & 87.84       & 76.95       & 80.89       & 62.54       & 73.80         \\
                                             & \sys              & 70.33       & 61.52       & 83.12       & 78.05       & 87.10       & 76.85       & 83.54       & 66.97       & 75.94         \\ \hline
      \multirow{4}{*}{Mixtral-Instruct-8x7B} & Matching          & 77.67       & 34.76       & 67.20       & 60.09       & 82.26       & 53.57       & 72.99       & 50.57       & 62.39         \\
                                             & Comparing         & 67.81       & 25.20       & 81.48       & 75.54       & 75.15       & 54.05       & 73.93       & 41.22       & 61.80         \\
                                             & Selecting         & 79.58       & 61.16       & 85.05       & 79.37       & 90.34       & 77.15       & 81.23       & 78.84       & 79.09         \\
                                             & \sys              & 84.13       & 72.51       & 87.32       & 82.03       & 92.33       & 81.67       & 83.82       & 82.48       & 83.29         \\ \hline
      \multirow{4}{*}{Solar-Instruct-10.7B}  & Matching          & 68.80       & 45.60       & 47.02       & 38.32       & 70.35       & 40.49       & 75.18       & 70.57       & 57.04         \\
                                             & Comparing         & 86.22       & 49.14       & 84.70       & 75.16       & 61.68       & 32.57       & 77.49       & 74.41       & 67.67         \\
                                             & Selecting         & 74.27       & 62.05       & 74.93       & 65.50       & 79.56       & 59.68       & 73.96       & 74.89       & 70.60         \\
                                             & \sys              & 78.83       & 61.01       & 62.25       & 61.92       & 79.52       & 67.48       & 76.46       & 74.85       & 70.29         \\ \hline
      \multirow{4}{*}{Flan-T5-XXL (11B)}     & Matching          & 77.85       & 58.35       & 87.63       & 80.34       & 71.82       & 51.62       & 74.62       & 67.23       & 71.18         \\
                                             & Comparing         & 84.21       & 56.85       & 94.49       & 85.82       & 65.33       & 49.88       & 84.28       & 67.89       & 73.60         \\
                                             & Selecting         & 77.52       & 69.83       & 84.77       & 80.29       & 85.07       & 68.05       & 78.90       & 77.33       & 77.72         \\
                                             & \sys              & 80.23       & 72.29       & 84.81       & 82.18       & 79.59       & 71.16       & 79.77       & 78.05       & 78.51         \\ \hline
      \multirow{4}{*}{Flan-UL2 (20B)}        & Matching          & 83.39       & 52.73       & 81.97       & 67.53       & 82.35       & 40.56       & 70.88       & 74.07       & 69.19         \\
                                             & Comparing         & 88.09       & 64.52       & 94.81       & 88.26       & 71.43       & 39.51       & 83.66       & 80.66       & 76.37         \\
                                             & Selecting         & 80.34       & 71.82       & 84.00       & 80.57       & 84.09       & 65.70       & 80.99       & 71.94       & 77.43         \\
                                             & \sys              & 81.27       & 74.27       & 85.14       & 81.71       & 80.52       & 68.13       & 81.42       & 80.98       & 79.18         \\ \hline
      \multirow{4}{*}{Command-R-35B}         & Matching          & 49.87       & 32.87       & 47.87       & 44.46       & 91.45       & 69.69       & 63.14       & 36.81       & 54.52         \\
                                             & Comparing         & 72.31       & 51.27       & 76.82       & 65.91       & 90.91       & 77.00       & 86.09       & 57.24       & 72.20         \\
                                             & Selecting         & 78.16       & 65.52       & 83.67       & 79.54       & 85.26       & 75.33       & 79.06       & 80.58       & 78.39         \\
                                             & \sys              & 78.34       & 69.24       & 84.29       & 80.97       & 85.80       & 76.81       & 78.39       & 78.44       & 79.03         \\ \hline
      \multirow{4}{*}{Llama-3.1-8B-Instruct} & Matching          & 53.97       & 29.93       & 22.97       & 26.54       & 80.14       & 62.00       & 65.94       & 40.56       & 47.75         \\
                                             & Comparing         & 84.41       & 55.27       & 85.80       & 75.49       & 72.20       & 70.33       & 78.39       & 85.30       & 75.90         \\
                                             & Selecting         & 78.29       & 68.00       & 81.16       & 75.57       & 79.65       & 75.68       & 77.45       & 74.82       & 76.33         \\
                                             & \sys              & 80.86       & 70.86       & 84.57       & 81.14       & 85.01       & 79.34       & 79.53       & 80.06       & 80.17         \\ \hline
      \multirow{4}{*}{Qwen2-7B-Instruct}     & Matching          & 63.41       & 47.33       & 68.35       & 52.46       & 82.89       & 55.54       & 71.84       & 55.06       & 62.11         \\
                                             & Comparing         & 84.32       & 56.88       & 88.78       & 76.57       & 93.17       & 65.07       & 86.50       & 75.39       & 78.34         \\
                                             & Selecting         & 72.39       & 61.03       & 81.49       & 76.57       & 82.97       & 73.48       & 78.55       & 72.96       & 74.93         \\
                                             & \sys              & 82.46       & 70.69       & 86.68       & 82.68       & 88.26       & 79.22       & 80.06       & 79.88       & 81.24         \\ \hline
    \end{tabular}%
  }
  \caption{F1 score of open-source LLMs under different strategies and \sys.}
  \label{tab:open-llm}
\end{table*}


\end{document}